\documentclass[11pt]{article}

\usepackage{acl}

\usepackage{times}
\usepackage{amsmath}
\usepackage{amssymb}
\usepackage{algorithm}
\usepackage{algpseudocode}  % from algorithmicx
\usepackage{float}

\usepackage{comment}

\usepackage[T1]{fontenc}
\usepackage[utf8]{inputenc}

\usepackage{microtype}

\usepackage{inconsolata}

\usepackage{graphicx}
\usepackage{multirow}
\usepackage{booktabs}

\title{Lost in Translation?\\ Exploring the Shift in Grammatical Gender from Latin to Occitan}

\author{
  \textbf{Ahan Chatterjee\textsuperscript{1,2}},
  \textbf{Matthias Schöffel\textsuperscript{1,2}},\\
  \textbf{Matthias Aßenmacher\textsuperscript{2,3}},
  \textbf{Marinus Wiedner\textsuperscript{4}},
  \textbf{Esteban Garces Arias\textsuperscript{2,3}} 
\\
\\
  \textsuperscript{1}Bavarian Academy of Sciences (BAdW), Munich
  \textsuperscript{2}LMU Munich\\
  \textsuperscript{3}Munich Center for Machine Learning (MCML)
  \textsuperscript{4}University of Freiburg
\\
  \small{
    \textbf{Correspondence:} \href{mailto:ahan.chatterjee@badw.de}{ahan.chatterjee@badw.de}
  }
}

\begin{document}
\maketitle
\begin{abstract}
The diachronic evolution from Latin to the Romance languages involved a restructuring of the grammatical gender system from a tripartite configuration (masculine, feminine, neuter) to a bipartite one (masculine, feminine) in most Romance languages. In this work, we introduce an interpretable deep learning framework to investigate this phenomenon at both lexical and contextual levels. First, we show that conventional tokenization strategies are insufficiently robust for this low-resource historical setting, and that our proposed tokenizer improves performance over these baselines. At the lexical level, we evaluate the contribution of morphological features to gender prediction. At the contextual level, we quantify the contributions of different part-of-speech categories to grammatical gender prediction. Together, these analyses characterize the distribution of gender information between the lemma and its sentential context. We make our codebase, datasets, and results publicly available at \href{https://github.com/ahan-2000/Lost-in-Translation-}{https://github.com/ahan-2000/Lost-in-Translation-}.
\end{abstract}

\section{Introduction}
\begin{figure}[t]
  \includegraphics[width=\columnwidth]{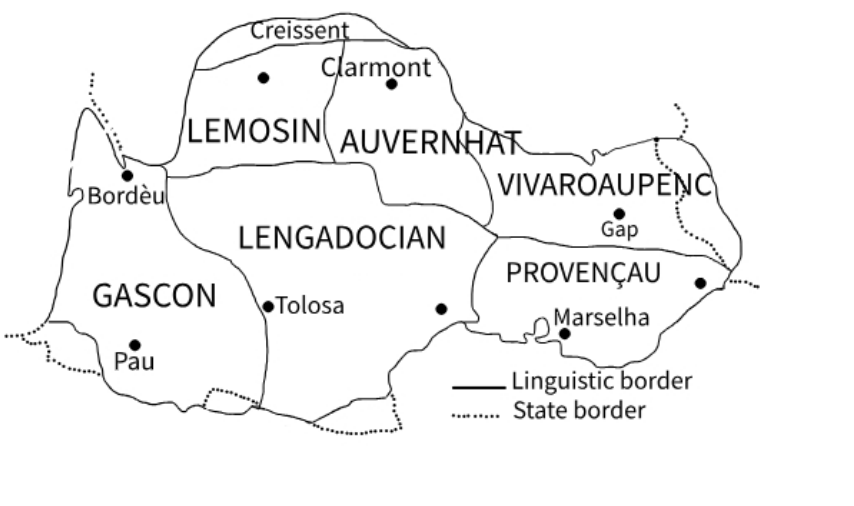}
  \caption{Historical spread of the Occitan language \cite{poujade-etal-2024-corpusarieja}.}
  \label{fig1:occitan map}
\end{figure}

Despite substantial advances in natural language processing (NLP), contemporary research remains concentrated on fewer than two dozen of the nearly 7,000 languages spoken worldwide. The vast majority of historical and regional languages are categorized as low-resource languages, defined by data scarcity, minimal digital presence, and a lack of standardized resources \citep{singh2008natural}. Medieval Occitan, a Romance language historically spoken in southern France, the Val d'Aran, and parts of Piedmont (cf. Figure \ref{fig1:occitan map}), played an important role in medieval cultural and economic life all over Europe. Despite this, UNESCO currently classifies it as an endangered language \citep{mothe2024shaping}. Medieval Occitan presents many of the challenges typical of low-resource languages. In addition to severe data scarcity and the lack of annotated gold-standard resources, the language displays substantial instability: it shows extensive orthographic variation, with lexical items attested in multiple spellings both across and within texts \citep{arias-etal-2023-automatic,schoffel2025unveiling, schoffel-etal-2025-modern}, as well as dialectal fragmentation stemming from the absence of a standardized norm \citep{zampieri2020natural}. As a result, existing work consistently describes Occitan as a neglected low-resource Romance language with severe resource limitations \citep{woller2021not}.

As a direct descendant of Vulgar Latin \citep{pasquini2021stability}, Occitan, too, underwent a transition from a tripartite to a bipartite gender system, as did most Romance languages. As Vulgar Latin evolved, morpho-phonological changes weakened the stable neuter category of Classical Latin, ultimately leading to the collapse and absorption of the neuters from the second declension class predominantly into the masculine gender \citep{szlovicsak2023preliminary}. However, the specific factors that governed this reassignment in Occitan, whether semantic, phonological, or morphological, remain insufficiently studied, especially for nouns inherited from the third declension class \citep{Marzo-Wiedner2025, polinsky2003development}. This work addresses this gap through a computational study of Medieval Occitan, examining how grammatical gender information is distributed between morphological features and morpho-syntactic context, and how these two sources contribute to model predictions for nouns descended from the Latin neuters.
Methodologically, we propose a general framework for disentangling lemma-internal and contextual signals in grammatical gender prediction for low-resource historical languages.
Our study is based on annotated corpora spanning law (\textit{Lo Codi}), medicine (\textit{Albuc}), and poetry (\textit{Croisade}). Using these heterogeneous but sparse resources, we examine how morphological features and morpho-syntactic context jointly contribute to gender prediction for nouns descended from the Latin neuter class, through the following research questions:

\paragraph{RQ1 (Lexical-Level Analysis):} To what extent can the grammatical gender of Occitan nouns derived from the Latin neuter class be predicted by word-level features, including their phonological and morphological characteristics?

\paragraph{RQ2 (Contextual Analysis):} How is grammatical gender information distributed between morphological features and morpho-syntactic context in Occitan, and how do these sources contribute to model predictions?

\section{Related Work}
During the medieval period, the collapse of Latin neuter nouns often led to their absorption into the masculine gender in Romance languages, i.e., for neuters from the second declension class ending in -\textit{um} \citep{klingebiel2019occitan, loporcaro2018gender}. Although morpho-phonological cues provide strong signals (e.g., nouns in \textit{-a} are typically feminine, while many others default to masculine), there are important exceptions, such as consonant-final gender-ambiguous nouns such as \textit{mar} (`sea'), as well as Grecisms in \textit{-a}, e.g., \textit{propheta} (`prophet'). These irregularities suggest that accurate gender assignment may require additional information, including stress patterns, Latin etyma, and, of course, sentence context, given that gender is not a morphological but a morpho-syntactic category and that Old Occitan lacks an overt gender system.

One core research question is how effectively grammatical gender can be assigned to a noun solely on the basis of its form, including its lexical, phonological, and morphological characteristics. Early work by \citet{brugmann1897nature} emphasized the critical role of both phonological and semantic cues in gender assignment. However, these approaches are largely rule-based and language-specific, limiting their generalizability across diverse languages or linguistic families. Classic typological research, such as \citet{Corbett_1991}, highlights that noun gender assignment typically involves a combination of morpho-phonological cues and semantic principles (e.g., natural gender for animates; but see the Greek loanwords as mentioned before). In Occitan, purely semantic gender applies in certain contexts, %—such as feminine agreement with the feminine referent ecologista—
but for inanimate nouns, formal phonological and morphological cues predominate in gender determination and sometimes even supersede semantic criteria, e.g. the Greek loanwords.

Computational studies have attempted to quantify and predict gender from lexical features. Rule-based approaches to gender assignment have been extensively developed for languages such as French, producing long lists of endings and their most probable genders \citep{lyster2006predictability}. \citet{nastase-popescu-2009-whats} analyze the prediction of grammatical gender using orthographic features and report that using a word's orthographic form in a statistical classifier improves gender prediction beyond baseline. These studies confirm that morpho-phonological cues have strong predictive power for gender. However, purely form-based prediction is not enough. Recent work by \citet{williams-etal-2019-quantifying} took an information-theoretic approach to languages such as German and Czech, measuring how much of gender assignment can be explained by a noun’s form, meaning, or inflection class. They found that a combination of features provides the best predictions, highlighting that no single feature (orthography, phonology, semantics) accounts for everything, which is supported by recent experimental evidence \citep{BasiratAllassonnièreTangBerdicevskis+2021}. Chronologically, the literature progressed from early descriptive grammars and implicit rules to manual rule compilations, then to data-driven classification, and now to neural and interpretable models. For Occitan, however, published computational work is still sparse. While some nouns have inherent grammatical gender, sentence context helps to identify gender assignment. In Occitan, as in related Romance languages like French and Spanish, determiners and adjectives agree in gender with nouns, participles, or pronouns. For example, the presence of the feminine article \textit{la} before a noun signals that the noun is feminine in that context. Thus, the noun \textit{torista} (‘visitor’) may be ambiguous in isolation, but in the phrase \textit{la torista}, the article disambiguates it as feminine in this context. While nouns predominantly have fixed grammatical gender, a few remnants, primarily from Latin neuter, exhibit atypical behaviour. Early computational work by \citet{cucerzan-yarowsky-2003-minimally} demonstrated that combining morphological analysis with contextual information significantly improves grammatical gender identification. Using a small annotated lexicon together with contextual cues such as co-occurrence with gendered articles and adjectives, their approach infers the gender of previously unseen words with high accuracy.

\section{Data Description}
The primary dataset for this study is drawn from three key Medieval Occitan sources. The first, \textit{Lo Codi}, was annotated by Tobias Schmid as part of the ALMA Project \citep{ALMA, prifti2023sprachdatenbasierte}. The second, the \textit{Chanson de la Croisade Albigeoise}, was prepared and revised by Marinus Wiedner. The third source is the DOM Dictionary Project \citep{DOM}. The resulting annotated dataset comprises Latin–Occitan pairs, including Latin words, their corresponding Occitan lemmata, and the grammatical gender of each form, with a data distribution of 40.85\% of unique lemmas from the \textit{DOM} data source, 46.39\% from \textit{Lo Codi}, and 12.76\% from \textit{Croisade}. In addition, we use raw Occitan texts to analyze contextual cues (cf. Appendix \ref{MATTR}).
\begin{figure}[ht]
  \centering
  \includegraphics[width=0.45\textwidth]{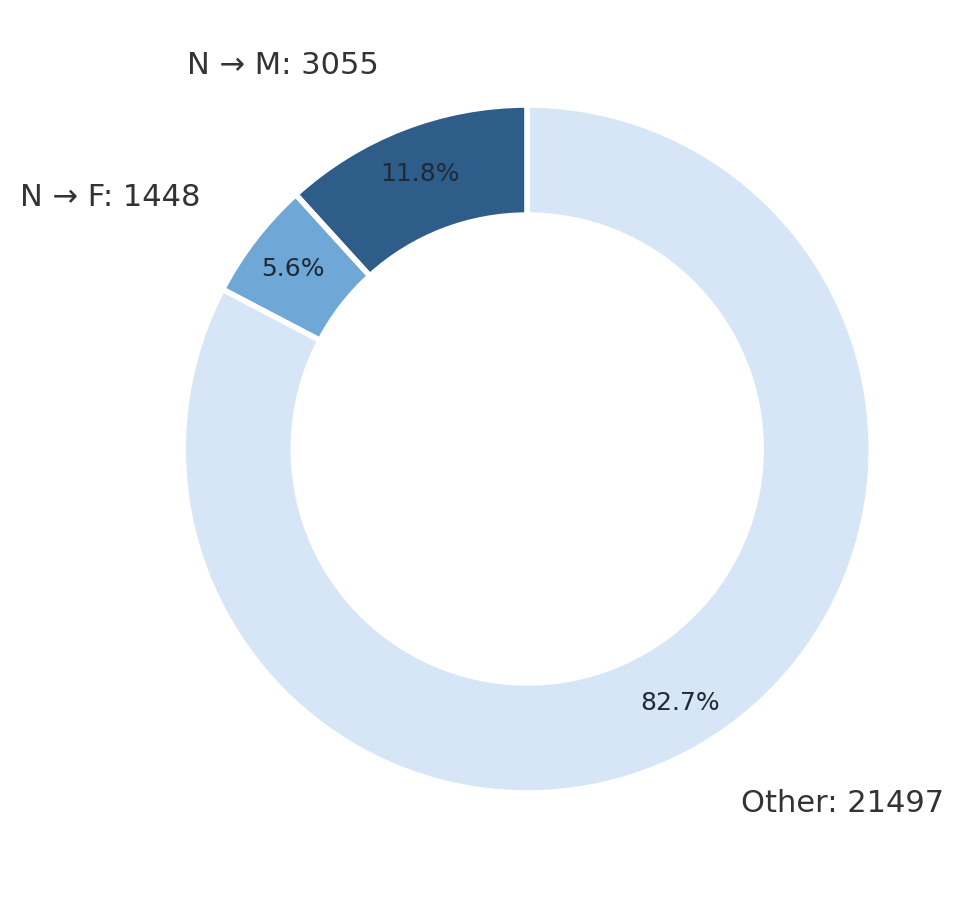}
  \caption {Gender Shift Frequencies across all three investigated corpora.}
  \label{fig:gender_frequencies}
\end{figure}

Our initial analysis confirms the complete absorption of the Latin neuter class into masculine and feminine genders in Occitan. As Figure \ref{fig:gender_frequencies} shows, the dominant shift is from neuter to masculine (3,055 cases), while a smaller but still substantial number of nouns shift to feminine (1,448 cases). A closer look at the orthographic features driving this divergence (Figure \ref{fig:gender_analysis_combined}, Appendix \ref{a:gender_lemma}) reveals that specific endings are highly predictive of the outcome. The role of endings in this process is more nuanced than raw frequency counts suggest. While the ending \textit{-um} is overwhelmingly associated with a masculine outcome, it is also, paradoxically, the single most common ending for nouns that become feminine. This is explained by the fact that the overall shift to masculine was far more prevalent, meaning any frequent neuter ending would appear dominant in that category. This finding underscores the importance of moving beyond simple frequency counts to understand the underlying mechanisms. By contrast, other endings, such as \textit{-ia} and \textit{-la}, provide a clearer signal and correlate strongly with feminine outcomes, further supporting the importance of morphological cues in gender (re)assignment.

\section{Preliminary Analysis: Model and Tokenization Selection}

We run a targeted set of probes to select (i) the embedding family that best captures Medieval Occitan variation in a Latin–Occitan setting and (ii) a tokenization policy that is robust to heavy orthographic noise. Concretely, we evaluate embedding models under three complementary criteria: (\(P_1\)) a frozen-encoder linear probe for Occitan gender prediction, (\(P_2\)) retrieval of Occitan orthographic variants given a Latin lemma, and (\(P_3\)) unsupervised structure in the embedding space via clustering.

\subsection{Embedding Model Selection}

We conduct a preliminary backbone selection study comparing FastText, mBERT, and ByT5 on three complementary probes: frozen gender prediction, Latin$\rightarrow$Occitan variant retrieval, and clustering of Occitan forms. mBERT performs best across all three probes, suggesting that it provides the most reliable representation of lexical and cross-lingual structure for Medieval Occitan. We therefore adopt it as the embedding backbone in all downstream experiments (cf. Appendix \ref{b:emb_model} for detailed results). 

\subsection{Tokenization Policy Selection}

Medieval Occitan exhibits frequent spelling variation and sparse type coverage, making segmentation a primary bottleneck. We therefore evaluate tokenization policies via (i) OOV rate and (ii) masked token recovery accuracy on an Occitan masked language modeling (MLM)-style objective.

\begin{figure}[h!]
\centering
\scriptsize
\setlength{\tabcolsep}{3pt}
\renewcommand{\arraystretch}{1.1}
\begin{tabular}{p{0.46\columnwidth} p{0.46\columnwidth}}
\textbf{Example 1} &
\textbf{Example 2} \\
\texttt{de lay del primpcipat} &
\texttt{En sa cambra secretament} \\
\emph{Hybrid:} \texttt{de, la, y, del, pri, mp, ci, pat} &
\emph{Hybrid:} \texttt{En, sa, cambra, s, ec, ret, amen, t} \\
\end{tabular}
\caption{Examples of hybrid tokenization capturing orthographic and morphological variation in Medieval Occitan. In \textit{primpcipat}, the subword \texttt{mp} isolates consonant-cluster variation, helping the model remain robust to spelling differences such as \texttt{nc}/\texttt{mp}. In \textit{secretament}, the final \texttt{t} is segmented separately, reflecting a common Old Occitan alternation where the adverbial \texttt{-t} may be elided (e.g., \textit{secretamen} vs.\ \textit{secretament}). Such fine-grained segmentation supports better generalization across predictable historical variants.}
\label{fig:tokenization_examples}
\end{figure}

\paragraph{Subword vs.\ Hybrid Segmentation.}
We evaluate tokenization policies using (i) \textbf{OOV rate}, defined as the proportion of tokens mapped to \texttt{[UNK]}, and (ii) \textbf{masked token recovery}, defined as top-1 accuracy at masked \emph{subword} positions. In this experiment, a hybrid policy (Occitan-adapted BPE with a word-level fallback) preserves full coverage (zero \texttt{[UNK]}) and yields the best masked recovery (25.23\%), indicating that explicit fallback coverage is crucial while still benefiting from corpus-adapted subword units (cf. Appendix \ref{tokenization_bpe}). Qualitatively, the hybrid tokenizer also produces more interpretable subword boundaries than generic WordPiece segmentation (Fig.~\ref{fig:tokenization_examples}).

\subsection{Domain-Adaptive MLM Fine-Tuning}

Given the consistent advantages of mBERT and hybrid tokenization, we apply domain-adaptive MLM fine-tuning for 10 epochs with identical hyperparameters across runs and evaluate on a held-out validation split. Fine-tuning substantially improves fit to the Occitan corpus: standard MLM adaptation reduces validation perplexity from 942.85 to 10.44, while the hybrid-vocabulary variant attains the best validation perplexity (9.52). Since perplexity is tokenization-dependent, we interpret these values as within-configuration diagnostics; taken together with the probing and tokenization results, they motivate our final setup: mBERT with a hybrid tokenizer and domain-adaptive MLM fine-tuning.

Based on the preliminary analyses, we adopt mBERT with hybrid tokenization and MLM adaptation as the backbone for all subsequent experiments. We now address our first research question by investigating grammatical gender prediction from lexical features alone, setting up a contrast with the contextual models introduced next. 

\section{Methodology}

\subsection{Lexical Grammatical Gender Prediction}
Grammatical gender is a nominal classification system; in Occitan, it is bipartite (masculine and feminine) and is typically realized through noun morphology and agreement. In this section, we examine gender assignment based solely on lexical information, without relying on sentential context.

\subsubsection{Feature Representation and Engineering}
\label{FeEngg}

\paragraph{Data Normalization.}
We lowercase lemmas and apply Unicode NFKD normalization, stripping combining diacritics for character-level features; original forms are retained for the embeddings.

\paragraph{Task and Imbalance Handling.}
We predict \emph{Occitan grammatical gender} as a bipartite label \(y\in\{\mathrm{M},\mathrm{F}\}\). Since outcomes are highly skewed (the Latin neuter most frequently maps to Occitan masculine), we use class-weighted training and focal loss; we also perform ablations to quantify the contribution of each feature group.

\paragraph{Morphological and Phonotactic Features.}
From both Latin and Occitan lemmas, we extract initial word substrings and suffix character \(n\)-grams (\(1\leq n\leq 4\)), emphasizing word-final cues consistent with Romance gender marking (Table~\ref{tab:bigram_example}). We further encode syllabic shape using (i) vowel-run syllable count \(S(w)\) and (ii) VC templates \(P(w)\) (Table~\ref{tab:syllabic_structure}), and include length features \( |w_{\text{lat}}|, |w_{\text{occ}}| \), their difference, and ratio.

\begin{table}[h!]
  \centering
  \resizebox{.4\textwidth}{!}{
  \begin{tabular}{llcc}
    \hline
    \textbf{Lang} & \textbf{Lemma} & \textbf{Initial Substrings} & \textbf{Suffix} \\
    \hline
    Latin   & \textit{domus} & \texttt{do} & \texttt{us} \\
    Occitan & \textit{dom}   & \texttt{do} & \texttt{om} \\
    \hline
  \end{tabular}
  }
  \caption{Example of initial word substrings/suffix bigram extraction (\(n=2\)) for aligned Latin--Occitan lemmas.}
  \label{tab:bigram_example}
\end{table}

\begin{table}[h!]
  \centering
  \resizebox{.4\textwidth}{!}{
  \begin{tabular}{llc}
    \toprule
    \textbf{Word} ($w$) & \textbf{Language} & \textbf{Syllable Count} $S(w)$ \\
    \midrule
    \textit{festum} & Latin   & 2 \\
    \textit{festa}  & Occitan & 2 \\
    \textit{tempus} & Latin   & 2 \\
    \textit{temps}  & Occitan & 1 \\
    \bottomrule
  \end{tabular}
  }
  \resizebox{.4\textwidth}{!}{
  \begin{tabular}{llc}
    \toprule
    \textbf{Word} ($w$) & \textbf{Language} & \textbf{VC Pattern} $P(w)$ \\
    \midrule
    \textit{festum} & Latin   & CVCCVC \\
    \textit{festa}  & Occitan & CVCCV  \\
    \textit{tempus} & Latin   & CVCCVC \\
    \textit{temps}  & Occitan & CVCCC  \\
    \bottomrule
  \end{tabular}
  }
  \caption{Syllabic structure features: vowel-run syllable count \(S(w)\) and VC template \(P(w)\).}
  \label{tab:syllabic_structure}
\end{table}

\paragraph{Stress as a Coarse Proxy.}
We include a lightweight stress-position proxy (ultimate/penultimate/antepenultimate), derived from a syllable-weight heuristic: monosyllables are stressed on the only syllable; disyllables on the penult; for polysyllables, we stress the penult if it is heavy (long vowel or closed syllable), otherwise the antepenult. We treat this feature as an approximate cue rather than as a definitive phonological annotation.

\paragraph{Embedding Features}
We use frozen pretrained representations as \emph{feature extractors} and compare them as alternative embedding feature sets rather than concatenating them: FastText (subword \(n\)-gram composition), mBERT, and ByT5. For mBERT/ByT5, each lemma is embedded in isolation and represented by mean pooling over subword/byte final-layer states; FastText uses standard word-type vectors. These embeddings are then used directly as input features to the downstream classifier.
\subsubsection{Experimental Setup}
We evaluate feature sets using lemma-grouped 10-fold cross-validation to prevent leakage across orthographic variants. Let \(\mathcal{D}=\{(x_i,y_i,\ell_i)\}_{i=1}^N\), where \(x_i\) are features, \(y_i\in\{\mathrm{M},\mathrm{F}\}\) is the label, and \(\ell_i\) is a lemma ID; folds are formed over lemmas and scores are averaged across folds. We evaluate a diverse set of classifiers to cover complementary inductive biases, ranging from transparent linear models (Logistic Regression), to non-linear tree ensembles (Random Forest, XGBoost), to sequence-aware neural architectures (FFN, BiLSTM, and attention-based variants). This design allows us to test whether grammatical gender is primarily recoverable from simple lexical cues or whether stronger performance requires models that capture higher-order or sequential interactions in the feature space. Hyperparameters are tuned with Optuna (Bayesian optimization), maximizing validation Macro-F1 within the cross-validation protocol.

Although lexical features are highly informative, they do not fully determine grammatical gender in all cases. For nouns such as \textit{psalmista}, the intended gender may only become clear from sentence-level agreement cues, especially the article (\textit{lo}/\textit{la}). We therefore turn to our second research question, examining whether contextual information improves prediction beyond lemma-internal evidence alone.

\subsection{Context-based Grammatical Gender Prediction}
In the previous section, we examined the contribution of lexical features to gender prediction in isolation. Here, we study the contribution of \emph{sentence-level context} as a second source. In Occitan, gender is jointly encoded by the noun and its agreeing dependents (articles, adjectives, and other modifiers); we exploit this distributed encoding as a prediction signal when lemma-internal cues are weak.

\subsubsection{Dataset \& Data Preparation}
We use $\sim$130k tokens of unannotated Occitan texts spanning multiple genres (law, poetry, and medicine). We normalize the corpus by lowercasing, stripping diacritics, and standardizing punctuation. Because parallel Latin sentences are unavailable, we rely on an existing Occitan--Latin lemma lexicon and link each Occitan lemma (cf. Algorithm \ref{alg:occ_lat_lemma_pipeline}) occurrence to its containing sentence, yielding contextual instances for downstream analysis.

\begin{algorithm}[t]
\caption{Construction of Occitan--Latin Lemma--Gender Dataset}
\label{alg:occ_lat_lemma_pipeline}
\footnotesize
\begin{algorithmic}[1]
\Require Raw Occitan corpus $D$
\Require Occitan--Latin lemma lexicon $\mathcal{L}$
\Require Similarity function $\textsc{Sim}(\cdot,\cdot)$ (cf. \ref{C: algo1})
\Ensure  Table $T$ of aligned lemmas, contexts and genders

\State $D_{\text{pos}} \gets \textsc{PoSTag}(D)$
\Comment{tag every token in the corpus (cf. \ref{PoS Appendix})}

\State \parbox[t]{\linewidth}{$N \gets \{(w,s,\ell_{\text{oc}})\in D_{\text{pos}} : \text{PoS}(w)=\textsc{NOUN}\}$}
\Statex \hspace{\algorithmicindent}\(\triangleright\) collect noun tokens with sentence $s$ and lemma $\ell_{\text{oc}}$

\State $T \gets \emptyset$

\ForAll{$(w,s,\ell_{\text{oc}}) \in N$}
    \Comment{iterate over all noun instances}

    \If{$\exists(\ell_{\text{oc}},\ell_{\text{la}},g_{\text{oc}},g_{\text{la}})\in\mathcal{L}$}
        \State $(\hat{\ell}_{\text{oc}},\hat{\ell}_{\text{la}},\hat{g}_{\text{oc}},\hat{g}_{\text{la}})
        \gets (\ell_{\text{oc}},\ell_{\text{la}},g_{\text{oc}},g_{\text{la}})$
        \Comment{exact lemma match}
        \Else
          \State Find $(\ell',\ell'_{\text{la}},g'_{\text{oc}},g'_{\text{la}})\in\mathcal{L}$ s.t.
          $\ell'=\arg\max_{\tilde{\ell}\in\mathcal{L}}\textsc{Sim}(\ell_{\text{oc}},\tilde{\ell})$
          and $\textsc{Sim}(\ell_{\text{oc}},\ell')\ge\tau$ \;($\tau{=}0.85$).

        \If{a candidate exists}
            \State $(\hat{\ell}_{\text{oc}},\hat{\ell}_{\text{la}},
            \hat{g}_{\text{oc}},\hat{g}_{\text{la}})
            \gets (\ell',\ell'_{\text{la}},g'_{\text{oc}},g'_{\text{la}})$
            \Comment{fuzzy lemma match}
        \Else
            \State \textbf{continue}
            \Comment{skip if no reliable match is found}
        \EndIf
    \EndIf

    \State \parbox[t]{0.30\linewidth}{Append row
    $(\hat{\ell}_{\text{oc}}, s, \hat{\ell}_{\text{la}}, \hat{g}_{\text{oc}}, \hat{g}_{\text{la}})$ to $T$}
    \Statex \hspace{\algorithmicindent}\(\triangleright\) store Occitan lemma, context, Latin lemma, and both genders
\EndFor

\State \Return $T$
\end{algorithmic}
\end{algorithm}

\subsubsection{Proposed Methodology}
We quantify the contribution of sentential context to Occitan gender prediction using three input settings. Each instance is \((X,i,L,G_L,y)\), where \(X=(x_1,\dots,x_T)\) is an Occitan sentence, \(i\) indexes the target noun token \(w=x_i\), \(L\) is its Latin lemma with Latin gender \(G_L\in\{\mathrm{M},\mathrm{F},\mathrm{N}\}\), and \(y\in\{\mathrm{M},\mathrm{F}\}\) is the gold Occitan label. A pretrained encoder produces contextual states:
\begin{figure}[ht]
  \includegraphics[width=\columnwidth]{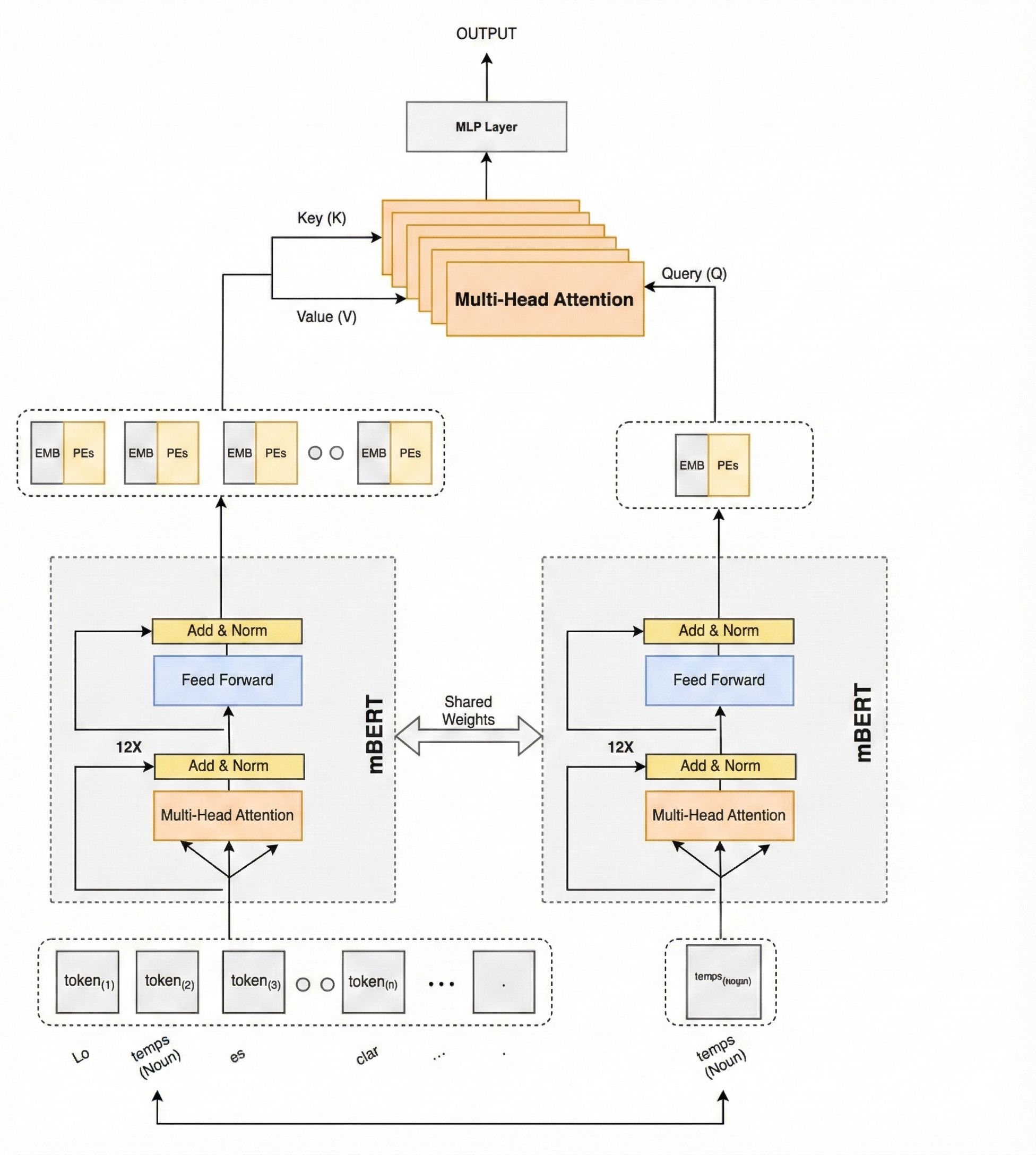}
  \caption{Proposed architecture to assess the impact of contextual cues on nouns' grammatical gender prediction.}
  \label{fig1:context arch}
\end{figure}
\begin{equation}
  \label{eq:ctx_encode}
  H=\mathrm{BERT}_\theta(X)=(h_1,\dots,h_T), \qquad h_t\in\mathbb{R}^d .
\end{equation}
All configurations share the same MLP head \(f_\phi\), with
\begin{equation}
  \label{eq:head}
  p(y\mid r)=\mathrm{softmax}\!\left(f_\phi(r)\right).
\end{equation}

\paragraph{(i) Word-only.}
We form a lexical representation from isolated embeddings and Latin metadata:
\begin{equation}
  \label{eq:r_wordonly}
  r_{\text{word}}=[\,e(w);\;e(L);\;\mathrm{onehot}(G_L)\,] .
\end{equation}

\paragraph{(ii) Context-focused.}
To target the noun within its sentence, we use noun-conditioned attention over \(H\) (cf. Architecture in Figure \ref{fig1:context arch}):
\begin{equation}
  \label{eq:r_ctx}
  r_{\text{ctx}}=[\,\mathrm{Attn}(h_i,H,H);\;e(L);\;\mathrm{onehot}(G_L)\,] .
\end{equation}

\paragraph{(iii) Masked-context.}
To isolate contextual cues, we mask the noun \(x_i\!\leftarrow\![\mathrm{MASK}]\), re-encode \(H^{\text{mask}}=\mathrm{BERT}_\theta(X^{\text{mask}})\), and use the state at position \(i\):
\begin{equation}
  \label{eq:r_mask}
  r_{\text{mask}}=[\,h^{\text{mask}}_i;\;e(L);\;\mathrm{onehot}(G_L)\,] .
\end{equation}

The masked-context setting evaluates how much of a noun's gender can be recovered from the surrounding sentence alone. Because Occitan articles, adjectives, and other dependents inflect to agree with the noun, the surrounding sentence jointly encodes the noun's gender; we therefore read masked-context performance as a measure of this distributed encoding rather than as an independent contextual signal. Comparing the word-only, context-focused, and masked-context configurations lets us bound how much predictive signal each source contributes (cf. Algorithm~\ref{alg:context_induction}).

All experiments use 3-group k-fold cross-validation, preserving the class distribution of Occitan gender labels across splits. To prevent label leakage, splits are constructed at the lemma level so that orthographic variants of the same lemma do not appear in both training and validation folds. A fixed random seed (13) is used throughout for reproducibility. We further analyze which contextual categories drive predictions by aggregating token-level contributions by PoS tag, e.g., determiners, adjectives, and verbs (cf. Appendix \ref{PoS Appendix}), yielding tag-wise estimates of their influence on gender prediction.

\section{Results and Discussion}

\subsection{Lemma-level gender prediction}
Table~\ref{tab:lemma-prediction-results} reports mean Accuracy and Macro-F1 (10-fold CV) for lemma-level gender prediction across model families and embedding feature sets. Overall, neural sequence models outperform shallow baselines, and attention generally yields further gains. The best Macro-F1 is achieved with a 2$\times$ BiLSTM + multi-head self-attention (MHSA) model trained with imbalance-aware objectives (focal loss/class weighting), yielding the strongest results with multilingual encoders (mBERT/ByT5), with pretrained representations providing more informative lexical cues than static embeddings.
\newpage

\begin{table*}[ht]
\centering
\small
\setlength{\tabcolsep}{3pt}
\renewcommand{\arraystretch}{1.05}
\resizebox{.75\textwidth}{!}{%
\begin{tabular}{lcc}
\hline
\textbf{Model} & \textbf{Accuracy} & \textbf{Macro F1} \\
\hline

\multicolumn{3}{c}{\textbf{ByT5}} \\
\hline
Logistic Regression & 0.6418 $\pm$ 0.0372 & 0.6139 $\pm$ 0.0440 \\
Random Forest & 0.7333 $\pm$ 0.0265 & 0.6974 $\pm$ 0.0333 \\
XGBoost & 0.7141 $\pm$ 0.0364 & 0.7354 $\pm$ 0.0492 \\
BiLSTM & 0.7550 $\pm$ 0.0410 & 0.7427 $\pm$ 0.0397 \\
$2\times$BiLSTM+Attn. (CE, 100 ep) & 0.7902 $\pm$ 0.0389 & 0.7867 $\pm$ 0.0354 \\
$2\times$BiLSTM+MHSA (CE, LS=0.1, 100 ep) & 0.8380 $\pm$ 0.0249 & 0.8106 $\pm$ 0.0237 \\
\hline

\multicolumn{3}{c}{\textbf{FastText}} \\
\hline
Logistic Regression & 0.6578 $\pm$ 0.0384 & 0.6319 $\pm$ 0.0459 \\
Random Forest & 0.7167 $\pm$ 0.0392 & 0.6917 $\pm$ 0.0397 \\
XGBoost & 0.7152 $\pm$ 0.0524 & 0.7068 $\pm$ 0.0528 \\
Feedforward NN (FFN) & 0.7266 $\pm$ 0.0231 & 0.7113 $\pm$ 0.0283 \\
BiLSTM & 0.7763 $\pm$ 0.0323 & 0.7561 $\pm$ 0.0293 \\
BiLSTM + Attn. (CE, 100 ep) & 0.7480 $\pm$ 0.0492 & 0.7224 $\pm$ 0.0536 \\
$2\times$BiLSTM+Attn. (CE+ES, 100 ep) & 0.7904 $\pm$ 0.0349 & 0.7512 $\pm$ 0.0354 \\
$2\times$BiLSTM+MHSA (CE+LS=0.1, 50 ep) & 0.8134 $\pm$ 0.0416 & 0.7734 $\pm$ 0.0344 \\
\hline

\multicolumn{3}{c}{\textbf{mBERT}} \\
\hline
Logistic Regression & 0.6749 $\pm$ 0.0254 & 0.6572 $\pm$ 0.0273 \\
Random Forest & 0.7287 $\pm$ 0.0397 & 0.7097 $\pm$ 0.0441 \\
XGBoost & 0.7352 $\pm$ 0.0402 & 0.7168 $\pm$ 0.0425 \\
BiLSTM & 0.7525 $\pm$ 0.0355 & 0.7252 $\pm$ 0.0384 \\
BiLSTM+Attn. (CE, 100 ep) & 0.7846 $\pm$ 0.0272 & 0.7419 $\pm$ 0.0309 \\
$2\times$BiLSTM+Attn. (FL + Class Wts, 50 ep) & 0.7840 $\pm$ 0.0369 & 0.7460 $\pm$ 0.0341 \\
\textbf{$2\times$BiLSTM + MHSA (CE, LS=0.1, 100 ep)} & \textbf{0.8327 $\pm$ 0.0365} & \textbf{0.8224 $\pm$ 0.0385} \\
\hline
\end{tabular}%
}
\caption{Mean $\pm$ SD over 10-fold lemma-grouped cross-validation for lemma-level grammatical gender prediction. Best per-embedding rows are bolded. Under the shared 2$\times$BiLSTM+MHSA head, the mBERT advantage over ByT5 is significant at the instance level on pooled out-of-fold predictions (paired bootstrap, $\Delta$ Macro-F1 $= +0.0395$, 95\% CI $[+0.0250,\,+0.0543]$, $p < 10^{-6}$; full procedure in Appendix~\ref{significane_mbert}).}
\label{tab:lemma-prediction-results}
\end{table*}

\begin{table}[H]
\centering
\small
\setlength{\tabcolsep}{4pt}
\scriptsize

  \resizebox{.45\textwidth}{!}{
\begin{tabular}{lccc}
\toprule
\multicolumn{4}{c}{\textbf{FastText (baseline = 0.7734)}} \\
\midrule
\textbf{Block} & \textbf{F1} & \(\Delta\) & \% drop \\
\midrule
Latin n-grams      & 0.7606 & 0.0128  & 1.66\% \\
Meta-features      & 0.7640 & 0.0094  & 1.22\% \\
Occitan n-grams    & 0.7667 & 0.0067  & 0.87\% \\
Syllable counts    & 0.7667 & 0.0067  & 0.87\% \\
VC patterns        & 0.7710 & 0.0024  & 0.31\% \\
Stress patterns    & 0.7746 & -0.0012 & -0.16\% \\
\bottomrule
\end{tabular}
}
\vspace{2pt}

  \resizebox{.45\textwidth}{!}{
\begin{tabular}{lccc}
\toprule
\multicolumn{4}{c}{\textbf{mBERT (baseline = 0.8224)}} \\
\midrule
\textbf{Block} & \textbf{F1} & \(\Delta\) & \% drop \\
\midrule
Latin n-grams      & 0.8092 & 0.0132  & 1.61\% \\
Meta-features      & 0.8168 & 0.0056  & 0.68\% \\
Occitan n-grams    & 0.8169 & 0.0055  & 0.67\% \\
Syllable counts    & 0.8194 & 0.0030  & 0.37\% \\
VC patterns        & 0.8220 & 0.0004  & 0.05\% \\
Stress patterns    & 0.8239 & -0.0015 & -0.18\% \\
\bottomrule
\end{tabular}
}
\vspace{2pt}

  \resizebox{.45\textwidth}{!}{
\begin{tabular}{lccc}
\toprule
\multicolumn{4}{c}{\textbf{ByT5 (baseline = 0.8106)}} \\
\midrule
\textbf{Block} & \textbf{F1} & \(\Delta\) & \% drop \\
\midrule
Latin n-grams      & 0.7958 & 0.0148  & 1.83\% \\
Meta-features      & 0.8006 & 0.0100  & 1.23\% \\
Occitan n-grams    & 0.8035 & 0.0071  & 0.88\% \\
Syllable counts    & 0.8087 & 0.0019  & 0.23\% \\
VC patterns        & 0.8091 & 0.0015  & 0.19\% \\
Stress patterns    & 0.8123 & -0.0017 & -0.21\% \\
\bottomrule
\end{tabular}
}
\caption{Feature ablation comparison across FastText, mBERT, and ByT5.}
\label{tab:ablation}
\end{table}
\subsection{Feature ablation}
To quantify the contribution of each feature group, Table~\ref{tab:ablation} removes one block at a time from the best configuration (per embedding set) and reports the resulting Macro-F1 drop. Latin and Occitan character \(n\)-grams, especially suffix cues, are the most influential, producing the largest decreases (1.6–1.8 Macro-F1 points). Length/meta-features are the next strongest contributors (0.7–1.3 points), while VC templates and stress proxies have comparatively smaller effects.

\subsection{Feature attributions (SHAP)}
\begin{figure}[H]
  \centering
\includegraphics[width=\columnwidth]{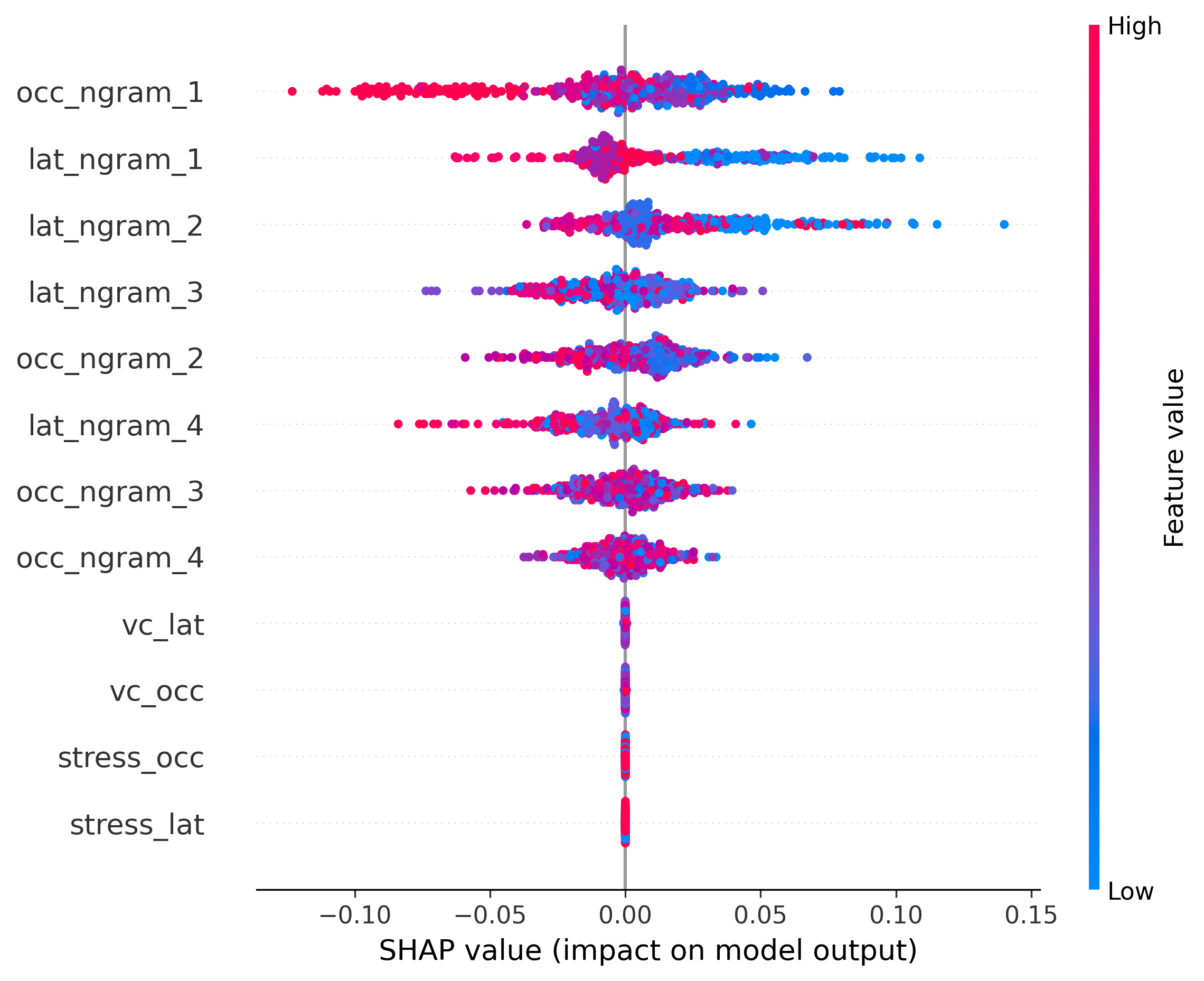}
  \caption{SHAP summary plot for the best-performing lemma-level model.}
  \label{fig:shap_placeholder}
\end{figure}

SHAP attributions (cf. Figure~\ref{fig:shap_placeholder}) are broadly consistent with the ablation results: suffix features and length-related meta-features dominate the decision signal across embedding modalities. Stress-related cues occasionally receive non-trivial attribution; however, since stress is derived from a heuristic proxy, we interpret these effects cautiously.

\begin{figure}[H]
  \centering
  \includegraphics[width=\columnwidth]{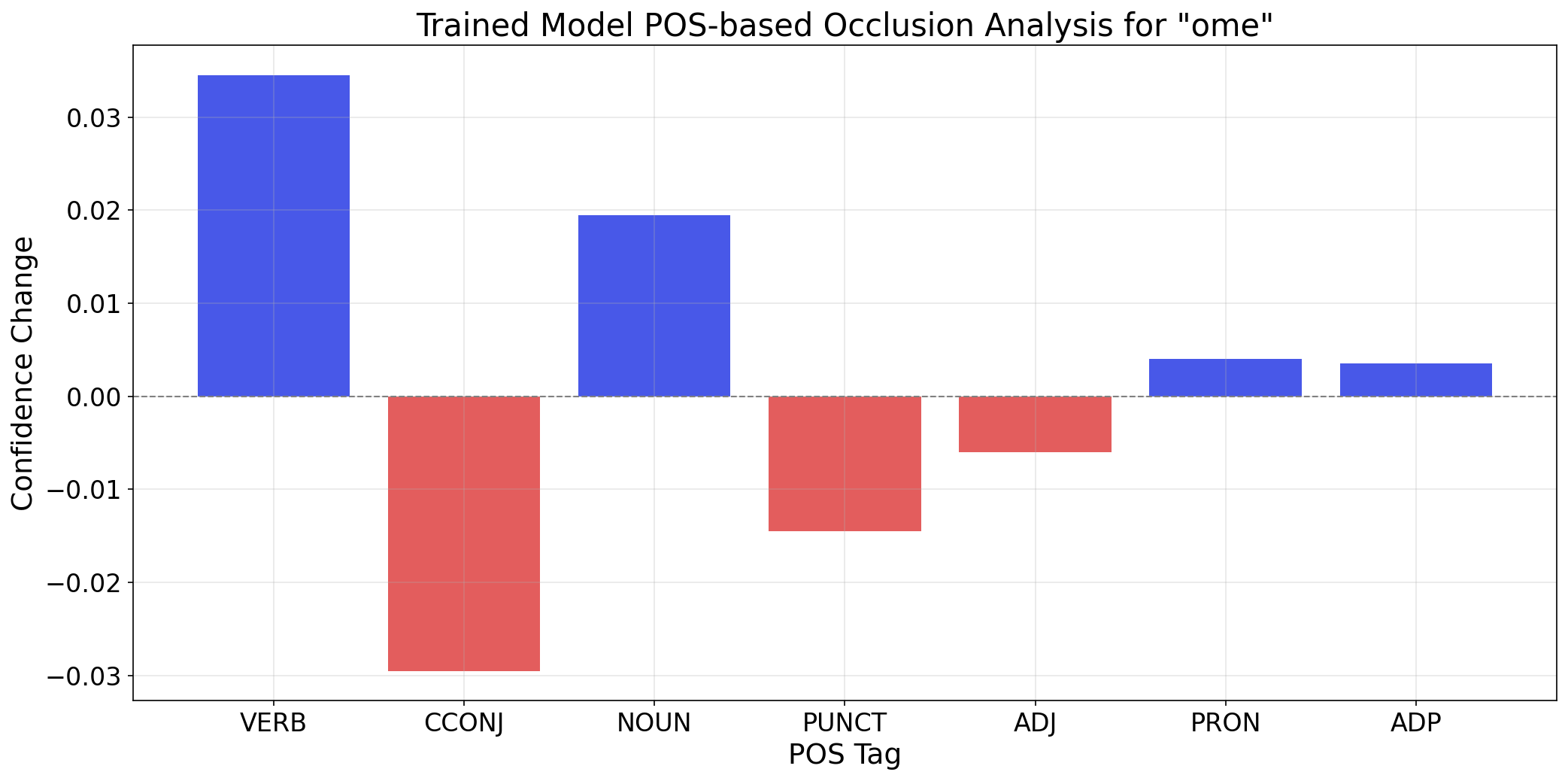}

  \caption{Example in which the lemma-only model misclassifies \textit{ome} as feminine: \textit{"aquill ome qui tenunt uera fe e pois tornunt en heresia deuent auer atrestal pena cum li altre e tant maior quant maior peccat ill fant."} With sentence-level context, the prediction shifts to masculine, with attribution distributed across agreement-bearing tokens such as \textit{aquill}, illustrating how gender information distributed between the lemma and its local context lets the contextual model recover the correct label when the lemma representation alone is insufficient.}  

  \label{fig:pos-imp-a}
\end{figure}

\subsection{Impact of Contextual Cues}
We evaluate contextual induction following Algorithm~\ref{alg:context_induction}. Table~\ref{tab:experiments} compares three mBERT-based configurations. Adding sentence context yields a substantial gain over the word-only baseline (Macro-F1: 0.665 \(\rightarrow\) 0.929). Masking the noun remains substantially better than word-only (Macro-F1 0.902), but underperforms the unmasked setting, consistent with the noun form carrying most gender signal while context provides additional disambiguation. In cases where the lemma representation alone is insufficient, the contextual model produces a different prediction than the lemma-only model, as illustrated in Figure~\ref{fig:pos-imp-a}. 

% Table: Experimental results (updated)
\begin{table}[H]
\centering
\small
\setlength{\tabcolsep}{4pt}
\renewcommand{\arraystretch}{1.05}
  \resizebox{\columnwidth}{!}{
\begin{tabular}{lcc}
\toprule
\textbf{Experiment (mBERT)} & \textbf{Acc.} & \textbf{Macro F$_1$} \\
\bottomrule
Word-only Model & 0.808 $\pm$ 0.154 & 0.665 $\pm$ 0.108 \\
\textbf{Context model (explicit noun attention)} & \textbf{0.979 $\pm$ 0.012} & \textbf{0.929 $\pm$ 0.034} \\
Context model (noun masked) & 0.977 $\pm$ 0.008 & 0.902 $\pm$ 0.097 \\
\bottomrule
\end{tabular}
}
\caption{Classification performance across the three experimental settings. The word-only baseline is lower than in the lemma-level experiments because, for comparability with the contextual models, it uses only the Latin lemma, Occitan lemma, and Latin gender, without the richer lemma-level feature set introduced earlier.}
\label{tab:experiments}
\end{table}

% Table: Delta statistics
\begin{table}[H]
\centering
\small
\setlength{\tabcolsep}{6pt}
\renewcommand{\arraystretch}{1.05}
  \resizebox{.4\textwidth}{!}{
\begin{tabular}{lcc}
\toprule
\textbf{$\Delta$ Statistic} & \textbf{Mean} & \textbf{95\% CI} \\
\midrule
$\Delta_{1}^{\text{prob}}$ & 0.283 & [0.281, 0.285] \\
$\Delta_{2}^{\text{prob}}$ & 0.279 & [0.277, 0.281] \\
$\Delta_{1}^{\log p}$      & 0.294 & [0.285, 0.303] \\
$\Delta_{2}^{\log p}$      & 0.340 & [0.334, 0.346] \\
\bottomrule
\end{tabular}
}
\caption{Mean values and 95\% confidence intervals for the $\Delta$ statistics.}
\label{tab:deltas}
\end{table}

To examine whether context increases confidence in the correct label, we report mean probability and log-probability deltas for the gold class (Table~\ref{tab:deltas}). Both \(\Delta_1\) (context vs.\ word-only) and \(\Delta_2\) (masked-context vs.\ word-only) are positive, indicating that contextual cues systematically raise the model’s confidence in the ground-truth class.

\subsection{Model Explainability}

For the context model, we use 8-head attention with the target noun state as query and sentence states as keys/values. Figure~\ref{fig:attn_combined} in Appendix \ref{app:exp} illustrates that attention concentrates on the noun token, with the associated article typically receiving the next-highest mass, matching Occitan morpho-syntax where articles (e.g., \textit{lo}/\textit{la}) are strong gender cues. Across heads, attention mass is broadly distributed, with no single head consistently specializing in a particular Part-of-Speech (PoS) category.

To quantify which contextual categories contribute most, we run PoS-conditioned occlusion (cf. Algorithm~\ref{alg:pos_impact} in Appendix \ref{algo3}) and aggregate token-level deltas by tag. Table~\ref{tab:pos-deltas} shows that nouns contribute the largest positive delta, followed by determiners and adjectives, consistent with gender information being distributed across the noun and its agreeing dependents.

\begin{table}[H]
\centering
\small
\setlength{\tabcolsep}{5pt}
\renewcommand{\arraystretch}{1.05}
  \resizebox{.45\textwidth}{!}{
\begin{tabular}{lccc}
\toprule
\textbf{PoS tag} & \textbf{Mean $\Delta$} & \textbf{Count (n)} & \textbf{Sign-flip $p$} \\
\midrule
NOUN  & $+0.0026$ & 39{,}521 & $<10^{-4}$ \\
DET   & $+0.0010$ & 24{,}042 & $<10^{-4}$ \\
ADJ   & $+0.0003$ & 29{,}920 & $<10^{-4}$ \\
CCONJ & $-0.0010$ & 27{,}492 & $<10^{-4}$ \\
ADP   & $-0.0007$ & 26{,}577 & $<10^{-4}$ \\
VERB  & $-0.0003$ & 29{,}336 & $<10^{-4}$ \\
PUNCT & $-0.0001$ & 30{,}408 & $0.096$ \\
PRON  & $-0.0002$ & 23{,}893 & $0.997$ \\
\bottomrule
\end{tabular}
}
\caption{PoS-wise mean occlusion deltas with sign-flip permutation test (10{,}000 permutations, two-sided). \textsc{Noun}, \textsc{Det}, and \textsc{Adj} contribute reliably positive contextual evidence; \textsc{Cconj}, \textsc{Adp}, and \textsc{Verb} contribute reliably negative effects; \textsc{Punct} and \textsc{Pron} are not significant. Effect magnitudes are small in absolute terms; we read them as stable but modest contextual cues. Full procedure in Appendix~\ref{significane_PoS}.}
\label{tab:pos-deltas}
\end{table}

\section{Conclusion}

Gender information in Medieval Occitan is distributed across two sources: lemma-internal morphology and sentence-level context. Suffix morphology carries the strongest single signal; articles, adjectives, and other agreeing dependents provide additional morpho-syntactic cues that may inform gender assignment in context-sensitive interpretation, and when the lemma alone is ambiguous, they can shift a model's prediction. Taken together, these findings support a two-layer view of gender in Medieval Occitan: lexical morphology provides the primary structural encoding, while agreement and contextual patterns, that is, morpho-syntactic cues, reflect its realization in usage. Methodologically, the work highlights that historical orthographic instability makes standard tokenization brittle; hybrid tokenization with domain-adaptive MLM enables models to exploit meaningful subword regularities while remaining robust to spelling variation. More broadly, the proposed lexical-versus-contextual comparisons and attribution analyses offer a useful framework for studying grammatical change in noisy historical corpora, though future work with richer gold annotation and improved morpho-syntactic resources would allow for more fine-grained analyses.

\section*{Limitations}

It is important to acknowledge several limitations. First, while our corpus is genre-diverse, it remains relatively small and label-imbalanced (approximately 2:1 masculine-to-feminine), which may limit minority-class generalization despite mitigation via focal loss and class-weighted training. Second, key components of the preprocessing pipeline were set heuristically, most notably the fuzzy-matching threshold (\(\tau=0.85\)) and the stress-position proxy, and our ablations suggest that the stress feature can introduce mild noise. Third, our PoS-conditioned analyses (cf. Appendix \ref{PoS Appendix}) rely on automatic tagging, and our evaluation shows \(\sim\)71\% tagging accuracy, implying that PoS-based attribution results may be biased by tagging errors. Finally, the contextual model is less reliable in sentences where the target noun occurs at sentence boundaries or where agreement-bearing cues (cf. Appendix \ref{error_analysis}) are sparse, which motivates future work on boundary-aware modeling and richer syntactic supervision. More broadly, our conclusions pertain to the Latin-to-Occitan neuter collapse and should be tested across additional Medieval Romance varieties. Our experiments quantify how gender information is \emph{distributed} between lexical and contextual sources for synchronic prediction; they do not directly test the diachronic question of what drove the historical reassignment of former Latin neuters, which requires parallel diachronic data and a different experimental design.

\section*{Ethics Statement}

We affirm that our research adheres to the 
\href{https://www.aclweb.org/portal/content/acl-code-ethics}{ACL Ethics Policy}. 
This work uses publicly available datasets and involves no human subjects or personally identifiable information. 
All data and code, including preprocessing, modeling choices, and evaluation protocols, are released to enable reproducible research and further investigation. 
%We acknowledge that language technologies may reflect biases present in the underlying training data and source materials, 
%which can influence downstream predictions and evaluations. 
Our work is intended exclusively for research purposes, and we encourage careful interpretation of results, 
particularly in low-resource and historical language settings where annotation uncertainty and data scarcity are common.

\section*{Acknowledgments}

Esteban Garces Arias sincerely thanks the Mentoring Program of the Faculty of Mathematics, Statistics, and Informatics at LMU Munich and the Munich Center for Machine Learning (MCML) for their ongoing support. Matthias Aßenmacher received funding from the Deutsche Forschungsgemeinschaft (DFG, German Research Foundation) under the National Research Data Infrastructure – NFDI 27/1 - 460037581 - BERD@NFDI.

\bibliography{custom}

\clearpage

\appendix

\onecolumn

\section{Data Description}
\subsection{Gender Shift by Lemma Ending}
\label{a:gender_lemma}

\begin{figure}[ht]
  \centering
  \includegraphics[width=0.82\linewidth]{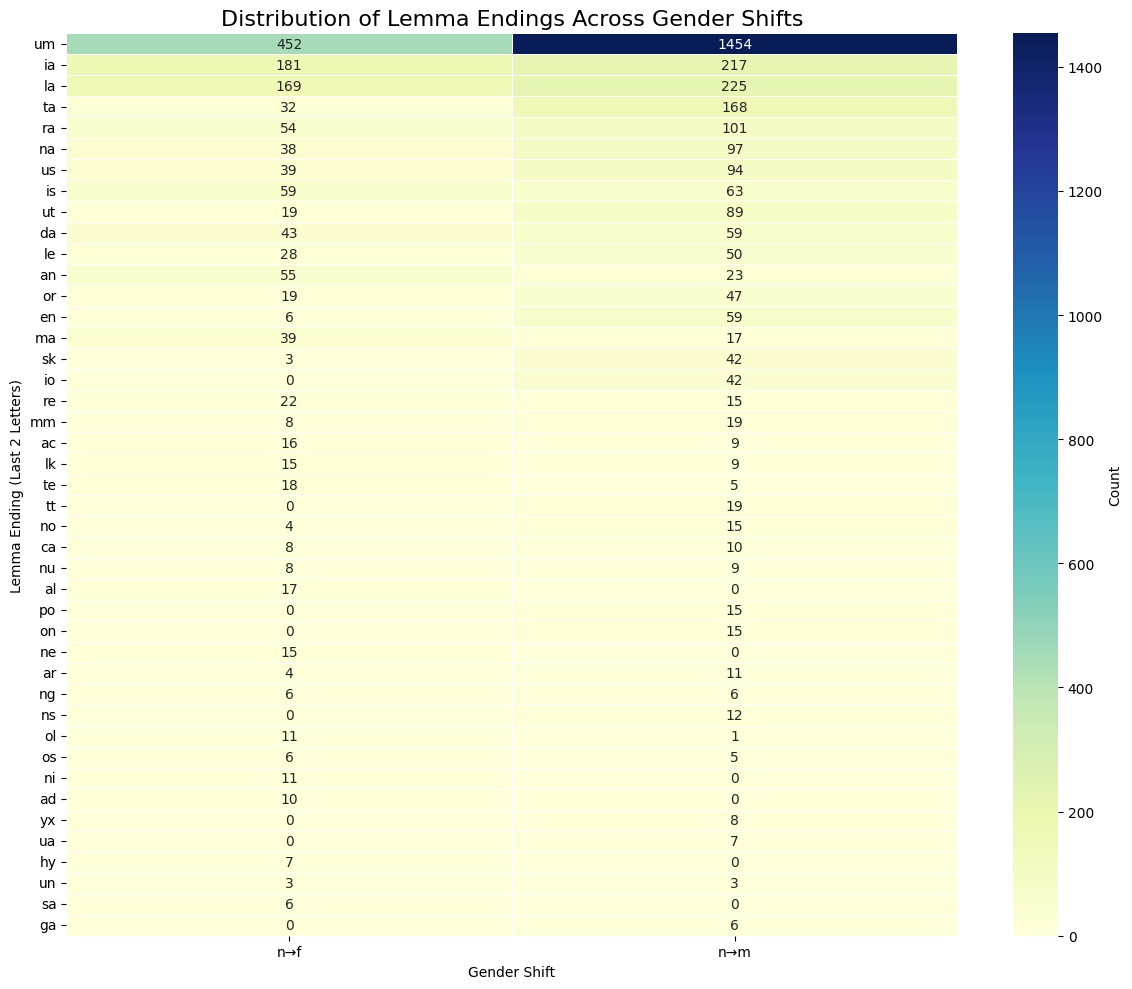}
  \caption {Gender shift frequencies for different lemma endings.}
  \label{fig:gender_analysis_combined}
\end{figure}
%\newpage
\subsection{Lexical Diversity in Raw Occitan Texts}
\label{MATTR}

\begin{table}[ht]
\centering
\resizebox{0.8\textwidth}{!}{%
\begin{tabular}{lrrrrrr}
\hline
\textbf{File} & \textbf{Tokens} & \textbf{Types} & \textbf{TTR} & \textbf{MATTR@50} & \textbf{MATTR@100} & \textbf{MATTR@500} \\
\hline
Harley\_3041.txt & 443 & 227 & 0.512 & 0.852 & 0.776 & 0.512 \\
Latin\_901.txt & 1105 & 492 & 0.445 & 0.830 & 0.739 & 0.537 \\
Nouvelle\_acquisition\_française\_11180.txt & 4312 & 1324 & 0.307 & 0.824 & 0.735 & 0.520 \\
Arsenal\_6355.txt & 11277 & 2491 & 0.221 & 0.849 & 0.762 & 0.545 \\
Français\_13504.txt & 35802 & 6163 & 0.172 & 0.808 & 0.724 & 0.536 \\
Roman\_de\_Flamenca.txt & 46724 & 7852 & 0.168 & 0.861 & 0.787 & 0.585 \\
Nouvelle\_acquisition\_française\_11151.txt & 20268 & 3151 & 0.155 & 0.766 & 0.661 & 0.429 \\
Croisade\_Albigeoise.txt & 81305 & 12370 & 0.152 & 0.822 & 0.747 & 0.562 \\
Add\_\_10323.txt & 53564 & 7609 & 0.142 & 0.850 & 0.764 & 0.552 \\
Add\_\_21218\_ohne1-6.txt & 31713 & 4485 & 0.141 & 0.805 & 0.717 & 0.507 \\
Français\_13503.txt & 35352 & 4773 & 0.135 & 0.820 & 0.722 & 0.493 \\
Français\_13509.txt & 62489 & 7867 & 0.126 & 0.849 & 0.763 & 0.548 \\
Français\_2232.txt & 34217 & 4194 & 0.123 & 0.799 & 0.705 & 0.489 \\
Lays\_d-amors.txt & 125475 & 13515 & 0.108 & 0.791 & 0.698 & 0.493 \\
Français\_1049.txt & 130220 & 10425 & 0.080 & 0.778 & 0.680 & 0.463 \\
\hline
\end{tabular}
}
\caption{Lexical diversity metrics for the 15 raw Occitan corpora, including the size-dependent TTR and the more robust MATTR metric, computed with varying window sizes (Data Source: \citet{wiedner2025cometa}).}
\end{table}
\newpage
\section{Preliminary Analysis}
\subsection{Embedding Model Backbone}
\label{b:emb_model}

\paragraph{(1) Frozen-encoder probing:}
We compare FastText \citep{bojanowski2016enriching}, mBERT \citep{devlin2019bert}, and ByT5 \citep{xue2022byt5} as frozen feature extractors with an identical linear classifier for Occitan \emph{grammatical gender} prediction. Each instance is a bilingual pair \((w_{\text{lat}}, w_{\text{occ}})\) with Latin gender \(g_{\text{lat}}\); we embed words in isolation, mean-pool subword/byte states, and classify \([e(w_{\text{lat}}); e(w_{\text{occ}}); \mathrm{onehot}(g_{\text{lat}})]\). mBERT performs best on this probe (\textbf{Macro F1 = 72.04}), outperforming FastText and slightly exceeding ByT5; we therefore adopt mBERT as our default backbone.

\begin{table}[h!]
\centering
\begin{tabular}{lcc}
\hline
\textbf{Embedding Model} & \textbf{Macro F1} & \textbf{Accuracy \%} \\
\hline
FastText & 57.51 & 58.30 \\
\textbf{mBERT} & \textbf{72.04} & \textbf{73.37} \\
ByT5 & 71.24 & 72.56 \\
\hline
\end{tabular}
\caption{Comparison of frozen embedding models on the Occitan gender prediction task.}
\label{tab:embedding_comparison}
\end{table}

\paragraph{(2) Variant retrieval:}
To test whether embeddings place one-to-many Latin $\rightarrow$ Occitan realizations near each other, we cast variant identification as retrieval: given a Latin lemma \(w_{\text{lat}}\), rank all candidate Occitan forms \(w_{\text{occ}}\) in the corpus by cosine similarity between isolated word embeddings (mean-pooled over subword/byte units). We report Recall@3 (and nDCG@3) since each query can have up to three attested variants in this experiment. Again, mBERT performs best on this probe (\textbf{Recall@3 = 0.59}), outperforming ByT5 and FastText, indicating that multilingual contextual encoders better cluster orthographic variants than static monolingual embeddings.
\begin{table}[h!]
\centering
\begin{tabular}{lccc}
\hline
\textbf{Embedding Model} & \textbf{Recall@k} & \textbf{MRR} & \textbf{nDCG} \\
\hline
FastText & 0.41 & 0.31 & 0.33 \\
\textbf{mBERT} & \textbf{0.59} & \textbf{0.47} & \textbf{0.49} \\
ByT5 & 0.51 & 0.40 & 0.42 \\
\hline
\end{tabular}
\caption{Retrieval performance for identifying Occitan orthographic variants from a Latin lemma.}
\label{tab:retrieval_performance}
\end{table}

\paragraph{(3) Unsupervised structure:}
We probe intrinsic geometry by applying K-Means to isolated Occitan form embeddings and evaluating cluster agreement with canonical lemma labels. mBERT yields the best cluster separation on this probe (\textbf{Silhouette = 0.049}), outperforming ByT5 and FastText; taken together, these findings motivate our use of mBERT in the downstream pipeline.
\begin{table}[h!]
\centering
\begin{tabular}{lcc}
\hline
\textbf{Embedding Model} & \textbf{Silhouette Score} & \\
\hline
FastText & 0.026 \\
\textbf{mBERT} & \textbf{0.049} \\
ByT5 & 0.042 \\
\hline
\end{tabular}
\caption{Clustering performance of different embedding models. Higher scores indicate better-defined and more pure clusters with respect to canonical lemmas.}
\label{tab:clustering_performance}
\end{table}

\subsection{Tokenization Policy on BPE}
\label{tokenization_bpe}

We compare the standard mBERT WordPiece tokenizer against corpus-trained BPE tokenizers (vocabulary sizes 600 and 800) and a hybrid tokenizer that combines Occitan-adapted BPE with a word-level fallback. Tokenizers are evaluated using two criteria: \textbf{OOV rate}, defined as the proportion of tokens mapped to \texttt{[UNK]}, and \textbf{masked token recovery}, defined as top-1 accuracy at masked subword positions.

\paragraph{BPE formulation.}
BPE iteratively builds a subword vocabulary by merging the most frequent adjacent pair of symbols. At step \(t\),
\begin{equation}
  \label{eq:bpe_merge}
  V_{t+1}=V_t \cup \{ab\}, \qquad (a,b)=\arg\max_{(x,y)} f(x,y),
\end{equation}
where \(f(x,y)\) is the corpus frequency of the pair \((x,y)\). Repeating this process for a fixed number of merges yields a vocabulary of reusable subword units.

\paragraph{Summary of findings.}
Table~\ref{tab:tokenization_performance} shows that corpus-trained BPE alone incurs non-zero OOV and very low masked recovery, while the standard mBERT tokenizer preserves full coverage but yields only moderate recovery. The hybrid tokenizer achieves the strongest overall trade-off, retaining zero OOV while substantially improving masked token recovery, which motivates its use in the downstream pipeline.

\begin{table}[ht]
\centering
\begin{tabular}{lcc}
\hline
\textbf{Tokenization Policy} & \textbf{OOV Rate (\%)} & \textbf{Masked Token Recovery (\%)} \\
\hline
mBERT Tokenizer & 0.0 & 15.78 \\
BPE (vocab=600) & 2.63 & 3.43 \\
BPE (vocab=800) & 2.86 & 4.76 \\
\textbf{Hybrid (BPE+word-level)} & \textbf{0.0} & \textbf{25.23} \\
\hline
\end{tabular}
\caption{OOV rate and masked token recovery for tokenization policies on the Occitan corpus.}
\label{tab:tokenization_performance}
\end{table}

\section{Algorithms}
\subsection{Algorithm 1: Construction of Occitan–Latin
Lemma–Gender Dataset}
\label{C: algo1}
We define
\[
\textsc{Sim}(x,y)
=
\alpha \,\textsc{CosSim}(x,y)
+
(1-\alpha)\,\textsc{LevSim}(x,y),
\]
with
\[
\textsc{LevSim}(x,y)
=
1-\frac{d_{\mathrm{Lev}}(x,y)}{\max(|x|,|y|)}.
\]
Since both \(\textsc{CosSim}(x,y)\) and \(\textsc{LevSim}(x,y)\) are normalized to \([0,1]\), \(\textsc{Sim}(x,y)\in[0,1]\). We set \(\alpha=0.3\), i.e.,
\[
\textsc{Sim}(x,y)
=
0.3\,\textsc{CosSim}(x,y)
+
0.7\,\textsc{LevSim}(x,y),
\]
and accept a candidate iff
\[
\textsc{Sim}(x,y)\ge 0.85.
\]
The threshold and the value for $\alpha$ were chosen through qualitative assessment across samples and threshold settings with an Occitan linguistic expert.
\newpage
\subsection{Algorithm 2: Evaluation of Contextual Induction in Grammatical Gender Prediction}
\label{algo2}
\begin{algorithm}[ht]
\caption{Evaluation of Contextual Induction in Grammatical Gender Prediction}
\label{alg:context_induction}
\footnotesize
\begin{algorithmic}[1]
\Require Dataset $D$ of input instances
\Require Models $M_\text{word}$ (word-only), $M_\text{ctx}$ (context), $M_\text{mask}$ (context with noun masked)
\Ensure \parbox[t]{0.9\linewidth}{Mean delta-probabilities and log-likelihood deltas for contextual induction; classification metrics}

\ForAll{sample $(X, i, W, L, G_L, Y)$ in $D$}
    \State $p_\text{word} \gets M_\text{word}(X, i, W, L, G_L)$
    \State $p_\text{ctx} \gets M_\text{ctx}(X, i, W, L, G_L)$
    \State $p_\text{mask} \gets M_\text{mask}(X, i, W, L, G_L)$
    \Statex \hspace{\algorithmicindent}\(\triangleright\) Ground-truth probability under word-only, context, and masked-context settings

    \State $\Delta_{p1} \gets p_\text{ctx} - p_\text{word}$
    \Comment{prob. deltas}
    \State $\Delta_{p2} \gets p_\text{mask} - p_\text{word}$

    \State $\Delta^{\log}_{p1} \gets \log p_\text{ctx} - \log p_\text{word}$
    \Comment{log-deltas}
    \State $\Delta^{\log}_{p2} \gets \log p_\text{mask} - \log p_\text{word}$

    \Statex \hspace{\algorithmicindent}\(\triangleright\) $\Delta_{p1}$: context vs word-only;\ \ $\Delta_{p2}$: masked-context vs word-only

    \State Record deltas and predicted labels for summary
\EndFor

\State Report $\text{mean}(\Delta_{p1})$, $\text{mean}(\Delta_{p2})$
\Comment{context induction (prob.)}
\State Report $\text{mean}(\Delta^{\log}_{p1})$, $\text{mean}(\Delta^{\log}_{p2})$
\Comment{context induction (log)}
\State \parbox[t]{0.9\linewidth}{Report accuracy and macro F1 for $M_\text{word}$, $M_\text{ctx}$, and $M_\text{mask}$}
\Comment{classification}
\end{algorithmic}
\end{algorithm}
%\newpage
\subsection{Algorithm 3: Estimating the Impact of PoS Tags on Grammatical Gender Prediction}
\label{algo3}
\begin{algorithm}[ht]
\caption{Estimating the Impact of PoS Tags on Grammatical Gender Prediction}
\label{alg:pos_impact}
\footnotesize
\begin{algorithmic}[1]
\Require Sentences $S$ with PoS tags (see Algorithm~\ref{alg:occ_lat_lemma_pipeline})
\Ensure \parbox[t]{0.9\linewidth}{Influence of PoS Tags}

\ForAll{sentence $s \in S$}
    \State Retrieve PoS tags $P = (p_1, p_2, \ldots, p_T)$ for $s$
        \Statex \(\triangleright\) e.g., attention-, gradient-, or
        perturbation-based scores

    \State Construct mapping between tokens and PoS tags
    \For{$t = 1$ to $T$}
        \State Mask token $x_t$ and recompute model confidence
        \Comment{occlusion}
        \State Record confidence change $\Delta c_t$
        \Comment{per token}
    \EndFor

    \State \parbox[t]{0.9\linewidth}{Aggregate token scores $(a_t)$ and/or confidence changes $(\Delta c_t)$ by PoS tag for sentence $s$}
\EndFor

\State Aggregate tag-wise statistics across all sentences
\State \Return PoS-tag contributions to gender prediction
\end{algorithmic}

\end{algorithm}

\newpage
\section{Model Architecture and Hyperparameters for the Experiments}
\label{model arch}
\subsection{Lemma Experiment}

\begin{table}[ht]
\centering
\small
\begin{tabular}{|l|l|p{0.65\textwidth}|}
\hline
\textbf{Embedding} & \textbf{Model} & \textbf{Best hyperparameters} \\
\hline
FastText & Logistic Regression
         & $C = 0.0027$; solver = \texttt{liblinear}; class\_weight = \texttt{None} \\
\hline
FastText & Random Forest
         & n\_estimators = 400; max\_depth = 23; min\_samples\_split = 6; min\_samples\_leaf = 5; class\_weight = \texttt{balanced} \\
\hline
FastText & XGBoost
         & n\_estimators = 100; max\_depth = 7; learning\_rate = 0.0210; subsample = 0.7344; colsample\_bytree = 0.8147; gamma = 0.6742; min\_child\_weight = 5 \\
\hline
FastText & Feedforward NN (FFN)
         & hidden\_dim = 64; dropout = 0.3934; lr = 0.0024; batch\_size = 16; optimizer = \texttt{Adam} \\
\hline
FastText & BiLSTM
         & hidden\_dim = 64; num\_layers = 1; dropout = 0.3552; lr = 0.00969; batch\_size = 16; optimizer = \texttt{Adam} \\
\hline
mBERT & Logistic Regression
         & $C = 0.0005$; solver = \texttt{liblinear}; class\_weight = \texttt{None} \\
\hline
mBERT & Random Forest
         & n\_estimators = 200; max\_depth = 41; min\_samples\_split = 8; min\_samples\_leaf = 5; class\_weight = \texttt{balanced} \\
\hline
mBERT & XGBoost
         & n\_estimators = 150; max\_depth = 9; learning\_rate = 0.0164; subsample = 0.5431; colsample\_bytree = 0.6104; gamma = 4.3585; min\_child\_weight = 5 \\
\hline
mBERT & BiLSTM
         & hidden\_dim = 256; num\_layers = 3; dropout = 0.1298; lr = 0.00025; batch\_size = 32; optimizer = \texttt{AdamW} \\
\hline
ByT5 & Logistic Regression
         & $C = 0.0733$; solver = \texttt{lbfgs}; class\_weight = \texttt{None} \\
\hline
ByT5 & Random Forest
         & n\_estimators = 400; max\_depth = 37; min\_samples\_split = 9; min\_samples\_leaf = 4; class\_weight = \texttt{None} \\
\hline
ByT5 & XGBoost
         & n\_estimators = 200; max\_depth = 3; learning\_rate = 0.0266; subsample = 0.8787; colsample\_bytree = 0.9173; gamma = 2.8813; min\_child\_weight = 2 \\
\hline
ByT5 & BiLSTM
         & hidden\_dim = 128; num\_layers = 1; dropout = 0.4417; lr = 0.00053; batch\_size = 32; optimizer = \texttt{AdamW} \\
\hline
\end{tabular}
\caption{Best hyperparameter settings for all models and embedding families.}
\label{tab:hyperparams}
\end{table}
\newpage
\subsection{Context-Level Experiment}
\label{arch context level}

\begin{table}[ht]
\centering
\resizebox{0.8\textwidth}{!}{%
\begin{tabular}{|l|c|c|c|}
\hline
\textbf{Hyperparameter} & \textbf{Exp1} & \textbf{Exp2} & \textbf{Exp3} \\
\hline
\multicolumn{4}{|l|}{\textit{Model Architecture}} \\
\hline
Model Type & FieldsOnlyModel & ContextReaderModel & ContextSimpleModel \\
\hline
Base Encoder & \multicolumn{3}{|c|}{BERT-base-multilingual-cased (768 dim)} \\
\hline
Encoder Frozen & Yes & No & No \\
\hline
Input Features & Word-level only & Context + Word & Masked Context + Word \\
\hline
Attention Mechanism & None & Multi-head (8 heads) & None \\
\hline
Attention Dimensions & N/A & $d_k=128$, $d_v=128$ & N/A \\
\hline
Relative Position Bias & N/A & Yes (window=64) & N/A \\
\hline
No Self-Peek & N/A & Yes & N/A \\
\hline
Feature Dimensions & 768 & 2560 & 2304 \\
\hline
MLP Hidden Size & 512 & 512 & 512 \\
\hline
\multicolumn{4}{|l|}{\textit{Training Hyperparameters}} \\
\hline
Epochs & 20 & 20 & 20 \\
\hline
Batch Size & 128 & 128 & 128 \\
\hline
Learning Rate & $2 \times 10^{-5}$ & $1 \times 10^{-5}$ & $2 \times 10^{-5}$ \\
\hline
Weight Decay & 0.01 & 0.01 & 0.01 \\
\hline
Warmup Ratio & 0.06 & 0.06 & 0.06 \\
\hline
Optimizer & \multicolumn{3}{|c|}{AdamW} \\
\hline
LR Schedule & \multicolumn{3}{|c|}{Linear warmup + linear decay} \\
\hline
Gradient Clipping & \multicolumn{3}{|c|}{0.5 (max norm)} \\
\hline
Dropout & 0.1 & 0.1 & 0.1 \\
\hline
Early Stopping Patience & \multicolumn{3}{|c|}{3 epochs} \\
\hline
Early Stopping Metric & \multicolumn{3}{|c|}{Validation loss} \\
\hline
Class Weights & \multicolumn{3}{|c|}{Balanced (sklearn)} \\
\hline
\multicolumn{4}{|l|}{\textit{Data Configuration}} \\
\hline
Max Sentence Length & \multicolumn{3}{|c|}{128 tokens} \\
\hline
Max Fields Length & \multicolumn{3}{|c|}{32 tokens} \\
\hline
Cross-Validation & \multicolumn{3}{|c|}{Group K-Fold (3 folds)} \\
\hline
Random Seed & \multicolumn{3}{|c|}{13} \\
\hline
\multicolumn{4}{|l|}{\textit{Experiment-Specific Details}} \\
\hline
Context Usage & None & Full sentence & Masked sentence \\
\hline
Masking Strategy & N/A & N/A & Noun replaced with \texttt{NOUNTOKEN} \\
\hline
Fields Format & \multicolumn{3}{|c|}{\texttt{[OCC]\{word\} [LAT]\{lemma\} [LG]\{gender\}}} \\
\hline
\end{tabular}%
}
\caption{Hyperparameters and training configuration for experiments 1, 2, and 3.}
\label{tab:hyperparameters}
\end{table}

\section{PoS Tagger in the Study}
\label{PoS Appendix}
The PoS tagger used in this study is \citep{manjavacas-etal-2019-improving}. A key limitation is that PoS tags are automatically predicted for the full Occitan corpus, and downstream analyses (including our occlusion-based PoS importance estimates) inherit tagging errors. To quantify tagger quality, we manually annotated a 60,000-token subset and evaluated the tagger against this gold data, obtaining 71.31\% overall accuracy. Performance varies by tag: \textsc{ADJ} shows the lowest accuracy, while our primary tag of interest, \textsc{NOUN}, achieves 70.32\%. We therefore interpret PoS-conditioned results as informative but potentially biased by tagging noise.

\section{Statistical Tests of the Experiments}
\subsection{Statistical Significance for Lemma Experiment}
\label{significane_mbert}
To complement the fold-averaged results in Table \ref{tab:lemma-prediction-results}, we directly compare mBERT and ByT5 under a matched downstream architecture using paired bootstrap resampling over \emph{out-of-fold} (OOF) predictions from the same 10 CV splits. Out-of-fold (OOF) predictions are obtained from the same lemma-grouped CV splits; therefore, each OOF prediction is made on a held-out lemma and is free of lemma-level leakage. This analysis is slightly different from Table \ref{tab:lemma-prediction-results}: there, Macro-F1 is reported as the mean across folds, whereas here we compute a single OOF Macro-F1 over all 4,444 held-out predictions to obtain a more robust paired comparison at the item level. Under the shared 2$\times$BiLSTM + MHSA classifier, mBERT yields a higher OOF Macro-F1 than ByT5 (0.7608 vs.\ 0.7213; \(\Delta=+0.0395\)). The paired bootstrap confirms that this advantage is reliable in the present setup: the 95\% confidence interval remains strictly above zero, and no bootstrap resample yields \(\Delta \leq 0\). 

\begin{table}[ht]
\centering
\begin{tabular}{lcc}
\hline
\textbf{System} & \textbf{OOF Macro-F1} & \textbf{Notes} \\
\hline
mBERT & 0.7608 &  \\
ByT5  & 0.7213 &  \\
\hline
\(\Delta\) (mBERT $-$ ByT5) & +0.0395 & 95\% CI [ +0.0250, +0.0543 ] \\
\(p\)-value & \multicolumn{2}{c}{\(p<10^{-6}\)} \\
\hline
\end{tabular}
\caption{Paired bootstrap comparison between mBERT and ByT5 under the same 2$\times$BiLSTM + MHSA architecture and identical 10-fold CV splits. Unlike Table~3, which reports mean Macro-F1 across folds, this table reports Macro-F1 computed over out-of-fold predictions for a paired item-level comparison. Because Table \ref{tab:lemma-prediction-results} reports fold means, we additionally perform a paired bootstrap on pooled out-of-fold predictions under the same architecture to test whether the mBERT–ByT5 difference is reliable at the instance level.}
\label{tab:bootstrap_mbert_byt5}
\end{table}

\subsection{Statistical Significance for PoS Tags}
\label{significane_PoS}

To assess whether the observed PoS-wise occlusion effects are reliably different from zero, we perform a \emph{sign-flip permutation test} for each PoS tag independently. For a given tag \(p\), let \(\delta_{i,p}\) denote the per-sample occlusion score, i.e., the change in confidence for the gold label when tokens with tag \(p\) are masked. Under the null hypothesis that the mean effect of \(p\) is zero, the sign of each \(\delta_{i,p}\) is exchangeable; we therefore generate a null distribution by randomly flipping the sign of the per-sample scores over 10,000 permutations and recomputing the mean. We report the observed mean effect together with the resulting 
$p$-value from a two-sided test of whether the mean effect differs from zero, which asks whether a PoS tag provides a reliably positive or negative contextual contribution. The full numerical results are reported in the main text in Table~\ref{tab:pos-deltas}.

\newpage

\begin{comment}
\begin{table}[ht]
\centering
\begin{tabular}{lccc}
\hline
\textbf{PoS tag} & \textbf{Mean \(\Delta\)} & \textbf{Count (n)} & \textbf{\(p\)-value} \\
\hline
NOUN  & +0.0026 & 39521 & \(<10^{-4}\) \\
CCONJ & -0.0010 & 27492 & \(<10^{-4}\) \\
DET   & +0.0010 & 24042 & \(<10^{-4}\) \\
ADP   & -0.0007 & 26577 & \(<10^{-4}\) \\
VERB  & -0.0003 & 29336 & \(<10^{-4}\) \\
ADJ   & +0.0003 & 29920 & \(<10^{-4}\) \\
PUNCT & -0.0001 & 30408 & 0.0957 \\
PRON  & -0.0002 & 23893 & 0.9966 \\
\hline
\end{tabular}
\caption{PoS-wise mean occlusion effects. \(p\)-values should be computed from sign-flip permutation tests applied to the same per-sample deltas used to obtain the reported means.}
\label{tab:pos_deltas_sig}
\end{table}
\end{comment}

\section{Explainability through Attention Heads on Contextual Cues Experiments}
\label{app:exp}
\begin{figure}[ht]
    \centering
    \includegraphics[width=\columnwidth,height=0.30\textheight,keepaspectratio]{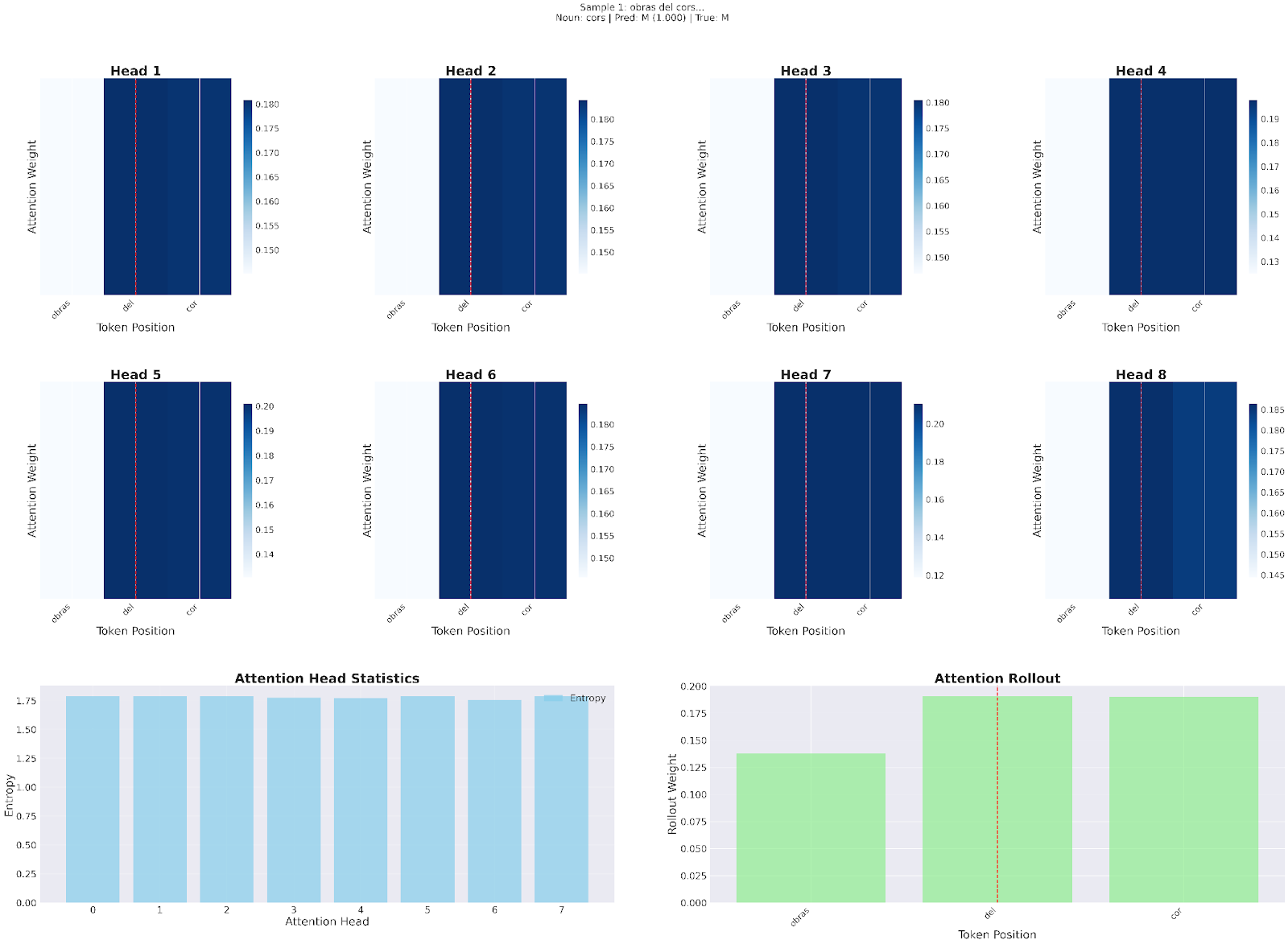}
    
    \vspace{0.4em}
    
    \includegraphics[width=\columnwidth,height=0.30\textheight,keepaspectratio]{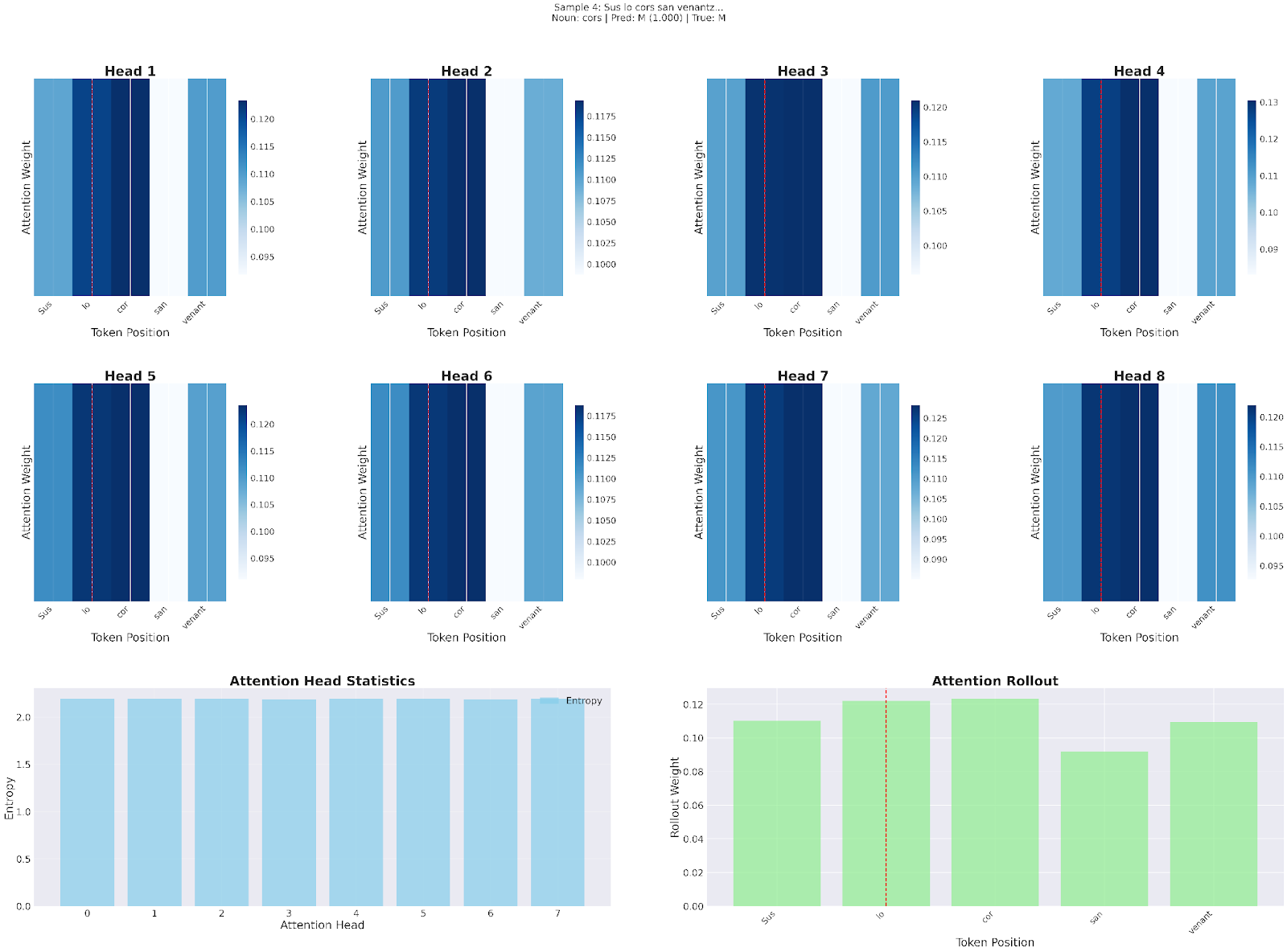}
    
    \caption{Attention-based contextual evidence for grammatical gender prediction shown for two representative Occitan sentences (top and bottom panels; target noun: \textit{cors}). For each example, we visualize the 8 MHSA heads when using the noun representation as the query and the full sentence as keys/values; the dashed red line marks the target noun position. Across heads, attention concentrates on the noun and nearby agreement-bearing tokens (often including determiners/articles), consistent with morpho-syntactic cues for gender assignment. The per-head entropy plot (left) indicates broadly distributed head behavior, and attention rollout (right) summarizes aggregate token attribution across the sentence.}
    \label{fig:attn_combined}
\end{figure}
\newpage
\section{Ablation Study on Contextual Cues Experiment}
\paragraph{Ablation: removing Latin features.}
To quantify the contribution of Latin etymological information in the contextual setting, we ablate the Latin lemma and Latin gender from the input and retrain/evaluate the same context model under 3-fold lemma-grouped cross-validation. Table~\ref{tab:ctx_ablation_nolatin} reports mean performance (± std across folds). While the context model remains effective without Latin features, the contextual \(\Delta\) gains in confidence are substantially reduced: with Latin lemma and gender, context increases the gold-class probability by \(\sim\)0.28 relative to the word-only baseline, whereas without Latin this increase drops to \(\sim\)0.09--0.11 (about 3$\times$ smaller). This indicates that Latin features provide critical complementary signal that amplifies the benefit of context.

\begin{table}[ht]
\centering
\begin{tabular}{lcc}
\hline
\textbf{Setting} & \textbf{Accuracy} & \textbf{Macro-F1} \\
\hline
Context + full noun (no Latin) & $0.961 \pm 0.018$ & $0.879 \pm 0.036$ \\
\hline
\end{tabular}
\caption{Contextual ablation without Latin lemma/gender. Results are mean $\pm$ std over 3-fold lemma-grouped cross-validation.}
\label{tab:ctx_ablation_nolatin}
\end{table}

\section{Error Analysis}
\label{error_analysis}
We analyse the 294 misclassifications made by the BiLSTM+Attention model across all folds using a SHAP-based surrogate approach. We train an XGBoost error predictor on 57 interpretable features capturing morphology (Latin/Occitan suffix cues, length, vowel ratio), frequency, sentence properties (length, noun position), and local syntax (PoS fractions and neighbouring tags) and explain its decisions with TreeSHAP (5-fold CV; ROC-AUC = 0.62). The strongest error drivers fall into three groups: (i) \emph{context sparsity}—sentences with fewer agreement-bearing categories, especially adjectives in the immediate right context, yield more errors; (ii) \emph{morphological ambiguity}—errors are more common when nouns occur at either sentence boundary; and (iii) \emph{length/frequency effects}—errors are more common in shorter contexts and for mid-frequency items, consistent with a regime where very frequent forms are memorised and rare regular forms generalise more easily. Overall, masculine items exhibit a higher error rate than feminine ones, though the difference is very low.

\begin{figure}[ht]
  \centering
  \includegraphics[width=0.79\linewidth]{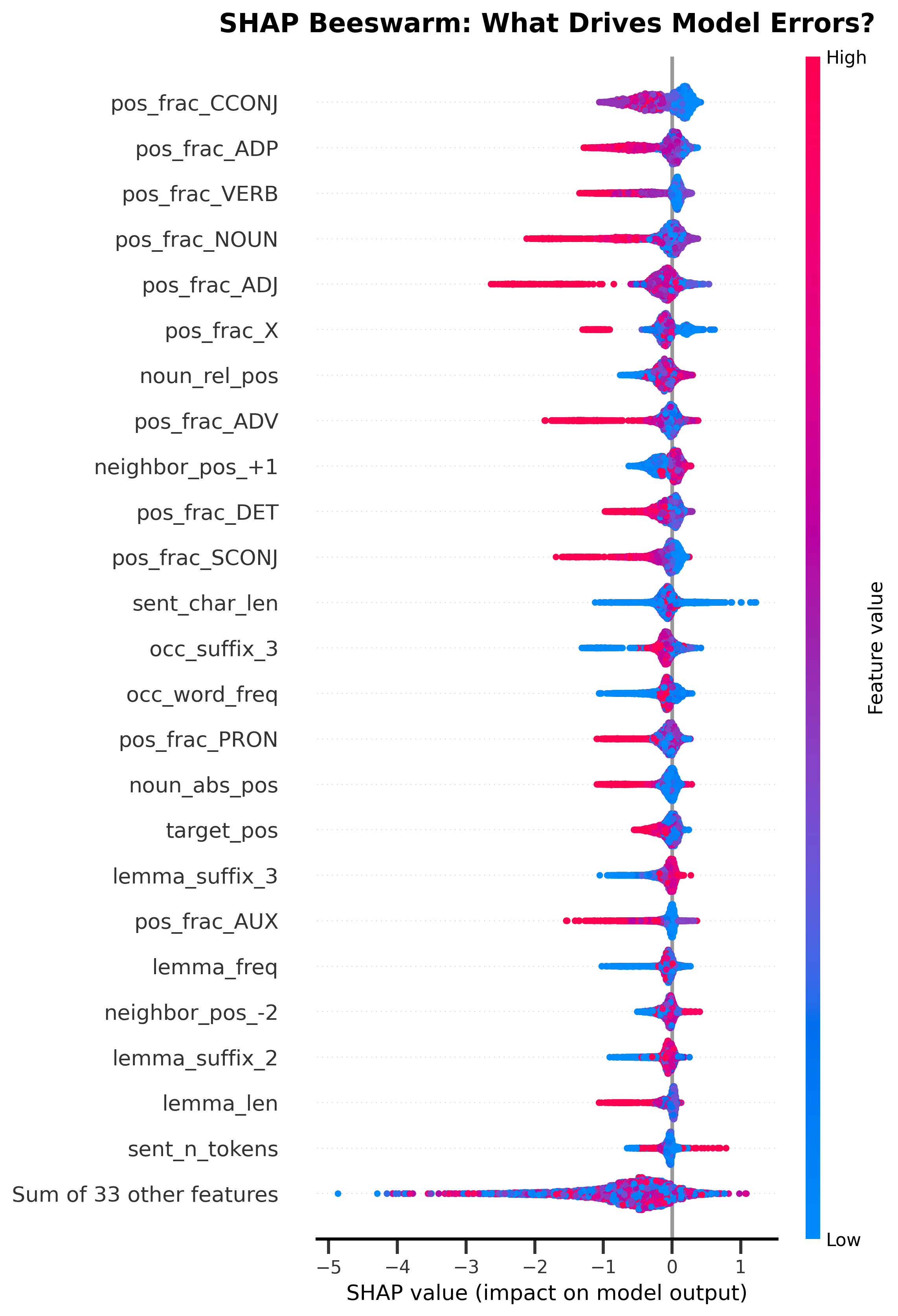}
  \caption {SHAP beeswarm plot showing feature contributions to model error prediction. Each dot represents a sample; the x-axis indicates the SHAP value (positive = pushes toward error, negative = toward correct), and colour encodes feature value (red = high, blue = low). The top five drivers are all POS composition features — fraction of coordinating conjunctions, adpositions, verbs, nouns, and adjectives in the sentence — indicating that syntactically sparse contexts lacking agreement-bearing words are the primary source of errors. Morphological features (Occitan suffix, Latin lemma suffix) and contextual factors (noun position, sentence length, word frequency) contribute secondarily. The immediate right-neighbour POS tag (rank 9) confirms that local agreement context, particularly an adjacent adjective, is a key disambiguating signal.}
  \label{fig:SHAP Plot of Error Analysis}
\end{figure}
\label{SHAP Error}

\end{document}